\documentclass[a4paper,11pt,twocolumn,twoside]{article}
\usepackage[dvips]{graphicx}
\usepackage{sepln_en}
\usepackage{fullname}
\usepackage[utf8]{inputenc}
\usepackage[spanish,es-nosectiondot, es-tabla, es-noindentfirst, es-nolists]{babel}

\input epsf

\title{Overview of GUA-SPA at IberLEF 2023: Guarani-Spanish Code Switching Analysis}

\author {\textbf{Luis Chiruzzo,$^1$} \textbf{Marvin Agüero-Torales,$^{2,3}$} \textbf{Gustavo Giménez-Lugo,$^4$} \textbf{Aldo Alvarez,$^5$} \\
\textbf{Yliana Rodríguez,$^1$} \textbf{Santiago Góngora,$^1$} \textbf{Thamar Solorio$^{6,7}$}\\
$^1$Universidad de la República, Montevideo, Uruguay\\
$^2$Universidad de Granada, Granada, Spain\\
$^3$Global CoE of Data Intelligence, Fujitsu, Madrid, Spain\\
$^4$Universidade Tecnologica Federal do Paraná, Curitiba, PR, Brazil\\
$^5$Universidad Nacional de Itapúa, Encarnación, Paraguay\\
$^6$University of Houston, Houston, TX, USA\\
$^7$Mohamed bin Zayed University of Artificial Intelligence, Abu Dhabi, United Arab Emirates\\
luischir@fing.edu.uy, maguero@correo.ugr.es, gustavogl@utfpr.edu.br,  aldo.alvarez@fiuni.edu.py,\\ yrodriguez@fhuce.edu.uy, sgongora@fing.edu.uy, thamar.solorio@gmail.com
}

\seplntranstitle{Overview de GUA-SPA en IberLEF 2023: Análisis de Cambio de Código en Guaraní-Español}

\seplnclave{Cambio de código, Guaraní, Español, NER.}

\seplnresumen{Presentamos la primera competencia de detección y análisis de cambio de código en guaraní y español, GUA-SPA en IberLEF 2023.
La competencia constó de tres tareas: identificar el idioma de cada token, NER, y una nueva tarea de clasificar la forma en que se usan los segmentos en español en el contexto del cambio de código.
Anotamos un corpus de 1500 textos extraídos de artículos de prensa y tweets, alrededor de 25 mil tokens, con la información necesaria para las tareas.
Tres equipos participaron de la fase de evaluación, obteniendo en general buenos resultados para la tarea 1, y resultados más dispares para las tareas 2 y 3.}

\seplnkey{Code-switching, Guarani, Spanish, NER.}

\seplnabstract{We present the first shared task for detecting and analyzing code-switching in Guarani and Spanish, GUA-SPA at IberLEF 2023.
The challenge consisted of three tasks: identifying the language of a token, NER, and a novel task of classifying the way a Spanish span is used in the code-switched context.
We annotated a corpus of 1500 texts extracted from news articles and tweets, around 25 thousand tokens, with the information for the tasks.
Three teams took part in the evaluation phase, obtaining in general good results for Task 1, and more mixed results for Tasks 2 and 3.}

\firstpageno{1}

\begin{document}

\renewcommand{\tablename}{Table}

% la siguiente instrucción sólo se debe usar si el abstract sobrescribe el texto
% la longitud variará según se necesite

%\setlength\titlebox{20cm} % se aumenta el tamaño del espacio reservado para datos de título

\label{firstpage} \maketitle

%\begin{abstract}
%Resumen del artículo con una sangría a izquierda y derecha de 0.32
%cm, justificado por ambos lados, con tamaño de fuente 11.
%
%\end{abstract}

\section{Introduction}

This paper presents the GUA-SPA shared task for code-switching analysis in Guarani and Spanish at IberLEF 2023~\cite{overviewIberLEF2023}.

Guarani is a South American indigenous language that belongs to the Tupi-Guarani family. Spanish is a Romance language that belongs to the Indo-European family, and both languages have been in contact in the South American region for about 500 years~\cite{rodriguez2018language}, resulting in many interesting linguistic varieties with different levels of mixture. 
Both, Guarani and Spanish, are Paraguay's official languages~\cite{ley4251}. 
According to the most recent census in Paraguay,\footnote{https://censo2022.ine.gov.py/noticias-actividades/21-de-febrero-dia-internacional-de-la-lengua-materna/} most of the country's population speaks at least some Guarani, and there is a high prevalence of Guarani-Spanish bilingualism in urban areas, while Guarani monolingualism is more limited to rural areas.

Bilingual speakers often make use of their two languages at the same time, mixing them in different ways, in a linguistic phenomenon called code-switching~\cite{joshi1982processing}. This phenomenon is very frequent in situations where two or more languages are in contact, for example, in linguistic borders, or in countries with large immigrant populations. In Paraguay, this has resulted in several identified language varieties that combine Guarani and Spanish~\cite{kallfell2016como}.
Two of the most common varieties of Guarani spoken in Paraguay are called Jopara and Jehe'a (both are Guarani words for describing a mix of things), and they often incorporate many Spanish loans of words or entire phrases, with varying levels of morphosyntactic adaptation to Guarani grammar~\cite{chiruzzo2022jojajovai}.

In this task, we propose to analyze a set of texts extracted from Paraguayan tweets and news articles, where it is usual to see these Jopara or Jehe'a varieties, and also the use of Spanish sentences that include Guarani loanwords.
These texts were annotated by the organisers to create the shared task dataset, which is available online.\footnote{https://github.com/pln-fing-udelar/gua-spa-2023} 
Using this dataset, we proposed three tasks: language identification (Task 1), NER (Task 2), and a novel classification task for the way Spanish spans are used in the code-switched context (Task 3).

The paper is structured as follows: in Section \ref{sec:background} we mention the previous work on code-switching analysis and Guarani processing; in Section \ref{sec:tasks} we detail the three tasks that make up this shared task; in Section \ref{sec:corpus} we present the corpus created for this work; in Section \ref{sec:competition} we present the competition, the systems built by the participant teams and their results; and in Section \ref{sec:conclusions} we detail the conclusions of this work.

\section{Background}
\label{sec:background}

There have been a number of competitions focusing on detection and analysis of code-switching, starting in~\namecite{solorio2014overview} with language identification in code-switched data for some language pairs, including Spanish-English. Later on these competitions started to include more complex tasks in code-switched contexts, such as NER~\cite{aguilar2020lince} and machine translation into the code-switched languages~\cite{chen2022calcs}.
As far as we know, this is the first time Guarani, or any other American indigenous language, is present as subject of a code-switching analysis competition, the first competition of NER in Guarani, and also the first dataset with this kind of annotations built for an American indigenous language.

However this is not the first work on Guarani identification.
\namecite{aguero2021logistical} tried to detect tweets in Guarani with heuristics using unigrams and bigrams and other out-of-the-box language identifiers that include Guarani, such as Polyglot \cite{al-rfou-etal-2013-polyglot}, fastText \cite{joulin2016bag} and NLTK's \cite{bird2009natural} implementation\footnote{https://www.nltk.org/\_modules/nltk/classify/textcat.html} of the method introduced by \namecite{Cavnar1994NgrambasedTC}.
\namecite{gongora2021experiments} used a Na\"ive Bayes model and a word-based heuristic to detect Guarani tweets.
Another big effort in language identification is HeLI-OTS \cite{jauhiainen-etal-2022-heli}, which indeed includes Guarani. 
Their authors were also participants of this shared task, as we will further describe in Section \ref{sec:systems_descriptions}.

Guarani is considered a low resource language~\cite{joshi2020state} because, despite having millions of speakers, it does not have many digital resources to work with, its written use online is scarce, and it has been mostly under-researched from the NLP perspective. The situation of this and other American indigenous languages could change in the future as there are now some initiatives to build resources for these languages~\cite{mager2021findings}, but there is still a long way to go.
Spanish, on the other hand, belongs to the small set of very resource-rich languages~\cite{joshi2020state}, which is good for this competition as there are many tools for Spanish that could be leveraged to see how well they work in this context.

\section{Tasks}
\label{sec:tasks}

The code-switched text detection and analysis tasks proposed in this challenge include the identification of the language used in each span of text, the named entities mentioned in the text, and the way Spanish is used. The challenge was structured as the following three tasks.

\subsection{Task 1: Language identification in code-switched data}

Given a text (sequence of tokens), label each token of the sequence with one of the following categories:

\begin{itemize}
    \item \texttt{gn}: It is a Guarani token.
    \item \texttt{es}: It is a Spanish token.
    \item \texttt{ne}: It is part of a named entity (either in Guarani or in Spanish).
    \item \texttt{mix}: The token is a mixture between Guarani and Spanish. For example a verb with a Spanish root that has been transformed into Guarani morphology, like: \textit{osuspendeta} (he/she will suspend).
    \item \texttt{foreign}: Used for tokens that are in languages other than Guarani or Spanish.
    \item \texttt{other}: Used for other types of tokens that are invariant to language, like punctuation, emojis and URLs.
\end{itemize}

Examples:

\begin{example}

\textit{che kuerai de pagar 6000 gs. por una llamada de 40 segundos . son aliados del gobierno parece ustedes}

Could be tagged as:

\textit{che/gn kuerai/gn de/es pagar/es 6000/other gs./other por/es una/es llamada/es de/es 40/other segundos/es ./other son/es aliados/es del/es gobierno/es parece/es ustedes/es}

\end{example}

\begin{example}

\textit{Ministerio de Salud omombe'u ko'ã káso malaria ojuhúva importado Guinea Ecuatorial guive ha oîma jesareko ohapejokóvo jeipyso .}

Could be tagged as:

\textit{Ministerio/ne de/ne Salud/ne omombe'u/gn ko'ã/gn káso/mix malaria/es ojuhúva/gn importado/es Guinea/ne Ecuatorial/ne guive/gn ha/gn oîma/gn jesareko/gn ohapejokóvo/gn jeipyso/gn ./other}

\end{example}

The metrics for Task 1 are accuracy, weighted precision, weighted recall and weighted F1. The main metric is weighted F1.

We implemented the following baseline for this task: choose the most frequent category for each word in the training corpus, or choose \texttt{other} if the word is not in the training corpus.

\subsection{Task 2: Named entity classification}

Given a text (sequence of tokens), identify the named entities as spans in the text, and classify each one with a category: person, location or organization.

Examples:

\begin{example}

\textit{[$_{ORG}$ Ministerio de Salud] omombe'u ko'ã káso malaria ojuhúva importado [$_{LOC}$ Guinea Ecuatorial] guive ha oîma jesareko ohapejokóvo jeipyso .}

\end{example}

\begin{example}

\textit{[$_{PER}$ Ministra de Hacienda Lea Giménez] he'i oñepromulga léi capitalidad ary 2014}

\end{example}

For this task, the tokens must be marked using the BIO notation, with the following labels: \texttt{ne-b-per}, \texttt{ne-b-loc}, \texttt{ne-b-org}, \texttt{ne-i-per}, \texttt{ne-i-loc}, \texttt{ne-i-org}.
The metrics for Task 2 are precision, recall and F1, either labeled or unlabeled. The criterion for finding a named entity is exact match. The main metric is labeled F1.

The baseline for this task is the following: take as an entity span any sequence of tokens labeled as \texttt{ne} by the baseline for Task 1, then choose the most frequent category (\texttt{per}, \texttt{org}, \texttt{loc}) for the first word of the sequence in the training corpus. If it is an unknown word, choose the category \texttt{per}.

\subsection{Task 3: Spanish code classification}

Given a text (sequence of tokens), identify spans of text in Spanish and label them in one of these categories:

\begin{itemize}
    \item \underline{change in code} (\texttt{CC}): the text keeps all the characteristics of Spanish.
    \item \underline{unadapted loan} (\texttt{UL}): the Spanish text could be partially adapted in some ways to Guarani syntax, but it is not fully merged into Guarani, in particular it does not present orthographic transformations.
\end{itemize}

Examples:

\begin{example}

\textit{che kuerai [$_{CC}$ de pagar 6000 gs. por una llamada de 40 segundos . son aliados del gobierno parece ustedes]}

\end{example}

\begin{example}

\textit{Okañývo pe Policía Nacional ha orekóva caso omomarandúvo Fiscalía , peteî [$_{UL}$ investigación ámbito penal] .}

\end{example}

For this task, the tokens must be marked using the following BIO labels: \texttt{es-b-cc}, \texttt{es-b-ul}, \texttt{es-i-cc}, \texttt{es-i-ul}.
The metrics for Task 3 are precision, recall and F1, either labeled or unlabeled. The criterion for finding a Spanish span is exact match. The main metric is labeled F1.

The baseline for this task is analogous to the one for Task 2: take as a Spanish span any sequence of tokens labeled as \texttt{es} by the baseline for Task 1, then choose the most frequent category (\texttt{cc}, \texttt{ul}) for the first word of the sequence in the training corpus. If it is an unknown word, choose the category \texttt{cc}.

\section{Corpus}
\label{sec:corpus}

The corpus of the task consists of 1500 texts, about 25 thousand tokens, split it in 76\%-12\%-12\% for train-dev-test.
The data contains sentences extracted from news articles~\cite{chiruzzo2022jojajovai,chiruzzo2020development} and tweets~\cite{aguero-et-al2023multi-affect-low-langs-grn,aguero2021logistical,gongora-etal-2022-use,rios2014sentiment}.
The characteristics of this corpus are shown in Table~\ref{tab:dataset}.
As seen in the table, the corpus contains more Guarani tokens than any of the other classes, for example there are on average three Guarani tokens per each Spanish token.
Although there are few mixed tokens, around 2\% of the corpus, they are a very interesting class to analyze as it presents particular challenges.

\begin{table} [h!]
\begin{center}
\begin{tabular} {|l|c|c|c|c|}
\hline\rule{-2pt}{15pt}
 & {\bf Train} & {\bf Dev} & {\bf Test} & {\bf Total} \\
\hline\rule{-4pt}{10pt}
Texts & 1140 & 180 & 180 & 1500 \\
\hline\rule{-4pt}{10pt}
gn & 7698 & 1241 & 1193 & 10132 \\
es & 5058 & 812 & 815 & 6685 \\
mix & 388 & 52 & 47 & 487 \\
ne & 2510 & 414 & 331 & 3255 \\
foreign & 129 & 14 & 8 & 151 \\
other & 3220 & 456 & 463 & 4139 \\
Total tokens & 19003 & 2989 & 2857 & 24849 \\
\hline\rule{-4pt}{10pt}
ne-per & 663 & 85 & 81 & 829 \\
ne-org & 562 & 95 & 89 & 746 \\
ne-loc & 240 & 58 & 33 & 331 \\
Total ne spans & 1465 & 238 & 203 & 1906 \\
\hline\rule{-4pt}{10pt}
es-ul & 636 & 80 & 95 & 811 \\
es-cc & 621 & 70 & 112 & 803 \\
Total es spans & 1257 & 150 & 207 & 1614 \\
  \hline
\end{tabular}
\end{center}
\caption{\label{tab:dataset}Composition of the dataset.}
\end{table}

The annotation was done in two phases:
A first phase consisted in four annotators annotating a sample of 54 sentences (around 750 tokens).
The aim of the first phase was to unify criteria and gauge how hard it was to make the annotations.
The four annotators obtained a Fleiss Kappa inter-annotator agreement of 0.836 for Task 1, which indicates very substantial agreement.
Only two of the annotators annotated the data for Task 2 during this phase, and we calculated their agreement using average F1, obtaining 0.926 between the two annotators, which we also consider a substantial agreement.
We came up with the idea for Task 3 during this pilot annotation, but it  was considerably harder to annotate, with many ambiguous cases that were difficult to generalize.

In the final phase of annotation, six annotators fluent in Spanish and with some knowledge of Guarani labeled data for Tasks 1 and 2, but only the three annotators with more knowledge of Guarani annotated the data for Task 3.

After the annotation was over we estimated the annotator agreement in Task 3 in the following way: we took a subsample of 90 texts that contained at least one token in Spanish, 30 for each annotator that participated in the task.
The three annotators completed the annotation of the 90 texts, and we compared the annotations as the F1 measure of one annotator against another.
The average F1 obtained in this way was 0.758 for the labeled F1 considering exact match which, although not as substantial as the agreement for Task 2, was still quite high.

\section{Competition}
\label{sec:competition}

The competition ran between March 22 and June 7 (2023) on the CodaLab platform\footnote{https://codalab.lisn.upsaclay.fr/competitions/11030}~\cite{codalab_competitions}. 
A total of 20 users registered to participate, but the number of participants that submitted results was considerably lower: three of them participated both in the development and the evaluation phase, and the other two participated in a single phase.

\subsection{Phases}

The competition consisted of three phases:\\

\noindent \textbf{Development phase}: from March 22 to May 23. 
This phase started with the publication of the training and development sets. 
During this phase the participants could submit their predictions for the development set and get the correspondent score for each task.
Each participant could make up to 200 submissions. 
There were 50 submissions.

\noindent \textbf{Evaluation phase}: from May 24 to June 6.
This phase started with the publication of the test set.
In this phase the participants could submit the predictions of their final systems and get the correspondent score for each task.
Each participant could make up to 10 submissions. 
There were 19 submissions.

\noindent \textbf{Post-Evaluation phase}: from June 7 onward.
This phase started after the end of the competition. 
The CodaLab page remains available for everyone who wants to test additional systems, download the training, development and test sets, and check the shared-task information.

\subsection{Systems Descriptions}
\label{sec:systems_descriptions}

Although we had only three participants in the evaluation phase, overall the followed approaches were notoriously diverse. 
We briefly describe those systems:

The user \textbf{aalgarra} submitted results only during the development phase, obtaining good results.
The chosen approach is based on finetuning a Spanish RoBERTa-base trained on BNE finetuned for CAPITEL NER dataset \cite{robertaBaseBNECapitelNER}, to which a bidirectional LSTM layer is added.
The target is obtained by assigning the category to the first token of each word and applying a mask in the loss calculation for the following tokens of each word.
The training was performed in two phases: an initial phase with frozen RoBERTa weights, with 30 epochs and a learning rate of $1e-3$, and a second phase training the full network during 2 epochs with a learning rate of $5e-5$.

The Universidade da Coruña team \cite{amunozoLySACoruna} (user \textbf{amunozo}) participated on all three tasks. 
They implemented their solution as a Multi-task learning problem where pre-trained encoder-decoder models were fine-tuned based on a hard parameter-sharing approach. 
They introduced three models with a common pre-trained LLM encoder fine-tuned on all three tasks and different decoders for each task. 
Their best-performing model on Task 1 is single-nllb model \cite{nllbteam2022language} where the encoder is trained only on Task 1 and the decoder is a softmax on the output layer, while the systems for Tasks 2 and 3 decode the output as Conditional Random Fields (CRF).
Their submitted systems use the mtl–beto-gn \cite{aguero-et-al2023multi-affect-low-langs-grn} model for Task 2 and the mtl-nllb model \cite{nllbteam2022language} for Task 3.
A post-processing heuristic is added to ensure the formatting of the final results. 

The University of Helsinki team \cite{tsjauhiaHeLI} (user \textbf{tsjauhia}) participated only on Task 1 obtaining the second place, very close to the winning team (0.0242 F1 points away).
They used a system consisting of their own HeLI-OTS language identifier \cite{jauhiainen-etal-2022-heli} based on the HeLI method \cite{jauhiainen-etal-2016-heli}, that includes Guarani.
Besides making some internal configurations to HeLI-OTS, they also added ad hoc rules, such as a text preprocessing pipline for Guarani and pattern-based logic to detect mixed Spanish-Guarani words.

\begin{table} [t!]
\begin{center}
\begin{tabular} {|c|c|c|c|}
  \hline\rule{-2pt}{15pt}
  {\bf User } & {\bf\small Task 1 } & {\bf\small Task 2 } & {\bf\small Task 3 } \\
              & {\bf\tiny Weighted F1 }   & {\bf\tiny Labeled F1 } & {\bf\tiny Labeled F1 } \\
  \hline\rule{-4pt}{10pt}
  pughrob & \textbf{0.9424} & \textbf{0.7604} & \textbf{0.4370} \\
  aalgarra & 0.9304 & 0.7047 & 0.2937 \\
  tsjauhia & 0.9053 & - & - \\
  amunozo & 0.8221 & 0.1986 & - \\
  \textit{baseline} & \textit{0.7032} & \textit{0.4218} & \textit{0.2331} \\
  \hline
\end{tabular}
\end{center}
\caption{\label{tab:dev_results}Results of the development phase.}
\end{table}

The Inclusive Technologies for Marginalised Languages (ITML) team \cite{pughrobITML} (user \textbf{pughrob}) from Indiana University obtained the first place on every task.
The authors evaluate a CRF model trained on text features and several neural network approaches using pre-trained multilingual representations, different fine-tuned BERT and T5 models.
The team got the first position in the competition for the three tasks, after finding that fine-tuning the multilingual representations with unlabeled monolingual Guarani data is beneficial for all three tasks, and that multi-task training achieves the best results for Task 2.

\subsection{Results}

Table \ref{tab:dev_results} shows the results of the development phase, while Table \ref{tab:test_results} shows the final results of the evaluation phase. 
We also had an additional submission by user \textbf{pakapro} for the evaluation phase that just tagged every token as \textit{other}, hence it is not included in the tables.

% The authors evaluate a CRF model trained on text features and several neural network approaches using pre-trained multilingual representations. They find that fine-tuning the multilingual representations with unlabeled monolingual Guarani data is beneficial for all three tasks, and that multi-task training achieves the best results for task 2
% They tried several approaches for the tasks, from a CRF model with classic text features to different fine-tuned BERT or T5 models. The team got the first position in the competition for the three tasks, using the fine-tuned BERT systems. Interestingly, the experiment that used the T5 model pretrained on American indigenous languages did not perform as well as expected.

\begin{table} [t!]
\begin{center}
\begin{tabular} {|c|c|c|c|}
  \hline\rule{-2pt}{15pt}
  {\bf User } & {\bf\small Task 1 } & {\bf\small Task 2 } & {\bf\small Task 3 } \\
              & {\bf\tiny Weighted F1 }   & {\bf\tiny Labeled F1 } & {\bf\tiny Labeled F1 } \\
  \hline\rule{-4pt}{10pt}
  pughrob & \textbf{0.9381} & \textbf{0.7028} & \textbf{0.3836} \\
  tsjauhia & 0.9139 & - & - \\
  amunozo & 0.8500 & 0.4153 & 0.1939 \\
  \textit{baseline} & \textit{0.7325} & \textit{0.4946} & \textit{0.2195} \\
  \hline
\end{tabular}
\end{center}
\caption{\label{tab:test_results}Results of the evaluation phase.}
\end{table}

In general, participants got very good results for Task 1, beating the baseline by a good margin.
The categories where the participants got the best performance for Task 1 were \texttt{gn}, \texttt{es}, and \texttt{other}, being the last one a very easy class to predict, containing mostly punctuation signs and URLs.
Conversely, the hardest classes to classify were \texttt{foreign} and \texttt{mix}, which also had the fewest examples.
The \texttt{mix} category is particularly interesting, as it comprises words that combine Guarani and Spanish content in some way, for example a Spanish verb with some adapted Guarani verbal morphology, or a Spanish noun with a Guarani case marker suffix.
There are only 47 examples of \texttt{mix} words in the test corpus, and three of those were correctly identified by all three participants: \textit{oñepresenta}, \textit{salud-pe}, and \textit{cartista-pe}.
The first one was in the training data, but not the others, so the systems were able to actually generalize at least to these simple cases.
Of the three participants, the University of Helsinki team was the only one that reported including particular rules for dealing with \texttt{mix} words by analyzing affixes, and they also got the best precision for this class, although the ITML achieved better recall.

The performance for Tasks 2 and 3 was more mixed, with only one team beating the baselines.
Task 3 was particularly hard for the automatic systems, and it was also hard for the annotators.
That can be seen as the best performance was only 0.384 for labeled F1.
Although it is not completely comparable, if we consider the annotators obtained an average F1 of 0.758 for a subsample of 90 texts, we can say that there is still a lot of room for improvement in this task.

\section{Conclusions}
\label{sec:conclusions}

In this work, we presented the GUA-SPA shared task for Guarani-Spanish code-switching analysis at IberLEF 2023, the first task of its kind that involves an American indigenous language.
The competition consisted of three tasks: language identification, NER, and a novel task of detecting the way Spanish is used in the code-switched context.
Three participants submitted their predictions for the evaluation phase, with the ITML team submission being the one with the best results across the three tasks.
Although performance for Task 1 was generally good for all participants, the results obtained for Tasks 2 and 3 were more mixed.
In particular Task 3, predicting if a Spanish span is used as an unadapted loan or as part of a code change, was the most challenging one both for the human annotators and the participants of the competition.
It would be interesting to explore this task more in the future, trying to better specify the two classes and how to deal with the ambiguous cases.
This could also be explored for other languages as well, as this phenomenon could be common in any code-switched context.
We would also like to incorporate more linguistic varieties of Guarani to the task, for example the Bolivian variety, and also include Portuguese as another possible language for the tokens, since it is another widely spoken language in South America, specifically near the areas where Guarani is spoken.
We hope this work contributes to spark the interest in code-switching analysis for more American indigenous languages.

% \section{Ejemplos de tablas y figuras}

% Ejemplo de tabla. En la Tabla \ref{tabla1} se muestran los
% resultados...
% \begin{table} [h]
% \begin{center}
% \begin{tabular} {|l|c|}
%   \hline\rule{-2pt}{15pt}
%   {\bf Descripción} & {\bf Cantidad}\\
%   \hline\rule{-4pt}{10pt}
%   Peras & 330\\
%   Manzanas & 70\\
%   Naranjas &  88\\
%   Limones & 16\\
%   Sandías & 73\\
%   \hline\rule{-2pt}{10pt}
%   {\bf Total}  & {\bf 577}\\
%   \hline
% \end{tabular}
% \end{center}
% \caption{\label{tabla1}Descripción.}
% \end{table}

% Ejemplo de Figura. En la Figura \ref{figura1} se muestran los
% resultados...

% \begin{figure}[h]
%   \centering
%   \includegraphics[width=5cm,clip]{ejem1.eps}
%   \caption{Descripción.}
%   \label{figura1}
% \end{figure}

% Este es un ejemplo de una referencia.

% Podemos referirnos a un autor poniendo su cita directamente
% \cite{Allen97}...

% También podemos referirnos a un autor incluyendo su nombre en la
% oración como ocurre con la propuesta de \namecite{allen2000}.

\bibliographystyle{fullname}
\bibliography{EjemploARTsepln}

\end{document}